\documentclass{article}
\usepackage{units}
\usepackage{booktabs} 
\usepackage[final]{corl_2017}
\usepackage[utf8]{inputenc}
\usepackage[T1]{fontenc}    
\usepackage[pdftex]{graphicx}
\usepackage{amsmath}
\usepackage{upgreek}
\usepackage{amsfonts,amssymb}
\usepackage{bm}
\usepackage{bbm}
\usepackage{amsthm} 
\usepackage{thmtools}
\usepackage{mathtools}
\usepackage{epstopdf}
\usepackage{xspace}
\usepackage{float}
\usepackage{lipsum}
\usepackage{soul}
\usepackage[font=footnotesize]{caption}
\usepackage[font+=small]{subcaption}
\usepackage[ruled,vlined,linesnumbered]{algorithm2e}  
\SetKwComment{Comment}{$\triangleright$\ }{}
\usepackage{enumitem}
\usepackage{tabulary}
\usepackage{multirow}
\newcommand{\algorithmStyle}[0]{\footnotesize}
\DeclareMathOperator*{\argmin}{arg\,min}

\newcommand{\pair}[2]{\left( #1, #2\right)}

\newcommand{\argminprob}[1]{\underset{#1}{\argmin}}

\newcommand{\abs}[1]{\left|#1 \right|}

\newcommand{\expect}[2]{\mathbb{E}_{#1}\left[#2\right]}
\newcommand{\prob}[1]{P\left(#1\right)}
\newcommand{\uniformSymb}[0]{\mathcal{U}}
\newcommand{\uniform}[1]{\uniformSymb\left(#1\right)}

\newcommand{\bbm}{\begin{bmatrix}}
\newcommand{\ebm}{\end{bmatrix}}

\newcommand{\set}[1]{\left\lbrace #1\right\rbrace}
\newcommand{\seq}[2]{\left(#1_{1}, #1_{2}, \ldots, #1_{#2}\right)}
\newcommand{\seqset}[2]{\left\lbrace#1_{1}, #1_{2}, \ldots, #1_{#2}\right\rbrace}

\newcommand{\Graph}[0]{G}
\newcommand{\vertexSet}[0]{V}
\newcommand{\edgeSet}[0]{E}

\newcommand{\vertex}[0]{v}
\newcommand{\edge}[0]{e}
\newcommand{\start}[0]{v_s}
\newcommand{\goal}[0]{v_g}
\newcommand{\succFnDef}[0]{\mathtt{Succ}}
\newcommand{\succFn}[1]{\succFnDef(#1)}
\newcommand{\Path}[0]{\xi}

\newcommand{\evalFn}[0]{\mathtt{Eval}}

\newcommand{\plannerTuple}[6]{\langle#1, #2, #3, #4, #5, #6\rangle}
\newcommand{\state}[0]{s}
\newcommand{\action}[0]{a}

\newcommand{\search}[0]{\mathtt{Search}}
\newcommand{\select}[0]{\mathtt{Select}}
\newcommand{\expand}[0]{\mathtt{Expand}}

\newcommand{\vertexSucc}[0]{V_\mathrm{succ}}
\newcommand{\invalidEdge}[0]{E_\mathrm{inv}}

\newcommand{\stateSpace}{\mathcal{S}}
\newcommand{\actionSpace}{\mathcal{A}}
\newcommand{\openList}[0]{\mathcal{O}}
\newcommand{\closedList}[0]{\mathcal{C}}
\newcommand{\closedObsList}[0]{\mathcal{I}}

\newcommand{\planTime}[0]{T}
\newcommand{\policy}[0]{\pi}
\newcommand{\policySpace}[0]{\Pi}
\newcommand{\policyOpt}[0]{\policy^*}
\newcommand{\policyLearn}[0]{\hat{\policy}}
\newcommand{\policyMix}[0]{\policy_\mathrm{mix}}
\newcommand{\policyOracle}[0]{\policy_\mathrm{OR}}
\newcommand{\policyOracleBel}[0]{\tilde{\policy}_\mathrm{OR}}

\newcommand{\costToGopolicy}[1]{Q_{t}^{#1}}
\newcommand{\costToGoOracle}{Q^{\textsc{COR}}}
\newcommand{\costToGo}{Q}

\newcommand{\world}[0]{\phi}

\newcommand{\worldSet}[0]{\mathcal{M}}
\newcommand{\worldProb}[0]{\prob{\world}}

\newcommand{\costonestep}[0]{c}
\newcommand{\cost}[0]{J}
\newcommand{\startProb}[0]{\prob{\start,\goal}}
\newcommand{\timeAvgDist}[0]{d_{\policy}^{t}}

\newcommand{\featureVec}{f}
\newcommand{\featureSpace}{\mathcal{F}}
\newcommand{\dataset}[0]{\mathcal{D}}
\newcommand{\thetaLearn}[0]{\hat{\theta}}
\newcommand{\thetaSpace}[0]{\Theta}
\newcommand{\mixParam}[0]{\beta}
\newcommand{\dataPointsParam}{k}
\newcommand{\errclass}[0]{\varepsilon_{\mathrm{class}}}
\newcommand{\errreg}[0]{\varepsilon_{\mathrm{reg}}}

\newcommand{\greedyEuc}{h_{\textsc{EUC}}}
\newcommand{\greedyMan}{h_{\textsc{MAN}}}
\newcommand{\distObs}{d_{OBS}}

\newcommand{\gVal}{g_{\vertex}}
\newcommand{\treeDepth}{d_{TREE}}
\newcommand{\bigo}[1]{\mathcal{O}\left(#1\right)}

\newcommand{\algName}[0]{\textsc{SaIL}\xspace}
\newcommand{\algFullName}[0]{Search as Imitation Learning \xspace}

\graphicspath{{figs/}}
\newcommand{\fullFigGap}[0]{\vspace{-1.5\baselineskip}}

\newtheorem{theorem}{Theorem}

\newtheoremstyle{hypstyle}
{3pt} 
{3pt} 
{\itshape} 
{} 
{\bfseries} 
{.} 
{.5em} 
{} 
\theoremstyle{definition}
\newtheorem{definition}{Definition}
\theoremstyle{hypstyle} 
\newtheorem{question}{Q}
\newtheorem{observation}{O}
\title{Learning Heuristic Search via Imitation\\}
\author{
  Mohak Bhardwaj\\
 The Robotics Institute\\
       Carnegie Mellon University\\
       \texttt{mohakbhardwaj@cmu.edu} \\
   \And
     Sanjiban Choudhury \\
     The Robotics Institute\\
     Carnegie Mellon University\\
     \texttt{sanjiban@cmu.edu} \\
   \And
   Sebastian Scherer \\
   The Robotics Institute\\
   Carnegie Mellon University\\
   \texttt{basti@cs.cmu.edu} \\
}
\begin{document}
\maketitle
\makeatletter
\renewcommand{\section}{%
\@startsection{section}{1}{\z@}%
            {-2.0ex \@plus -0.0ex \@minus -1.0ex}%
            { 1.5ex \@plus  0.0ex \@minus  0.5ex}%
            {\large\bf\raggedright}%
}
\providecommand{\subsection}{}
\renewcommand{\subsection}{%
  \@startsection{subsection}{2}{\z@}%
                {-1.2ex \@plus -0.0ex \@minus -0.8ex}%
                { 0.8ex \@plus  0.0ex }%
                {\normalsize\bf\raggedright}%
}
\providecommand{\subsubsection}{}
\renewcommand{\subsubsection}{%
  \@startsection{subsubsection}{3}{\z@}%
                {-0.8ex \@plus -0.0ex \@minus -0.5ex}%
                { 0.5ex \@plus  0.0ex}%
                {\normalsize\bf\raggedright}%
}
\makeatother
\setlength \abovedisplayskip{1pt plus0pt minus1pt}
\setlength \belowdisplayskip{\abovedisplayskip}

\begin{abstract}
Robotic motion planning problems are typically solved by constructing a search tree of valid maneuvers from a start to a goal configuration. 
Limited onboard computation and real-time planning constraints impose a limit on how large this search tree can grow.
Heuristics play a crucial role in such situations by guiding the search towards potentially good directions and consequently minimizing search effort.
Moreover, it must infer such directions in an efficient manner using only the information uncovered by the search up until that time. 
However, state of the art methods do not address the problem of computing a heuristic that \emph{explicitly minimizes search effort}.
In this paper, we do so by training a heuristic policy that maps the partial information from the search to decide which node of the search tree to expand. 
Unfortunately, naively training such policies leads to slow convergence and poor local minima. 
We present \algName, an efficient algorithm that trains heuristic policies by imitating \emph{clairvoyant oracles} - oracles that have full information about the world and demonstrate decisions that minimize search effort. We leverage the fact that such oracles can be efficiently computed using dynamic programming and derive performance guarantees for the learnt heuristic.
We validate the approach on a spectrum of environments which show that \algName consistently outperforms state of the art algorithms.
Our approach paves the way forward for learning heuristics that demonstrate an anytime nature - finding feasible solutions quickly and incrementally refining it over time.
\end{abstract}

\section{Introduction} 

\begin{figure}[!t]
    \centering
    \includegraphics[page=1,width=\textwidth]{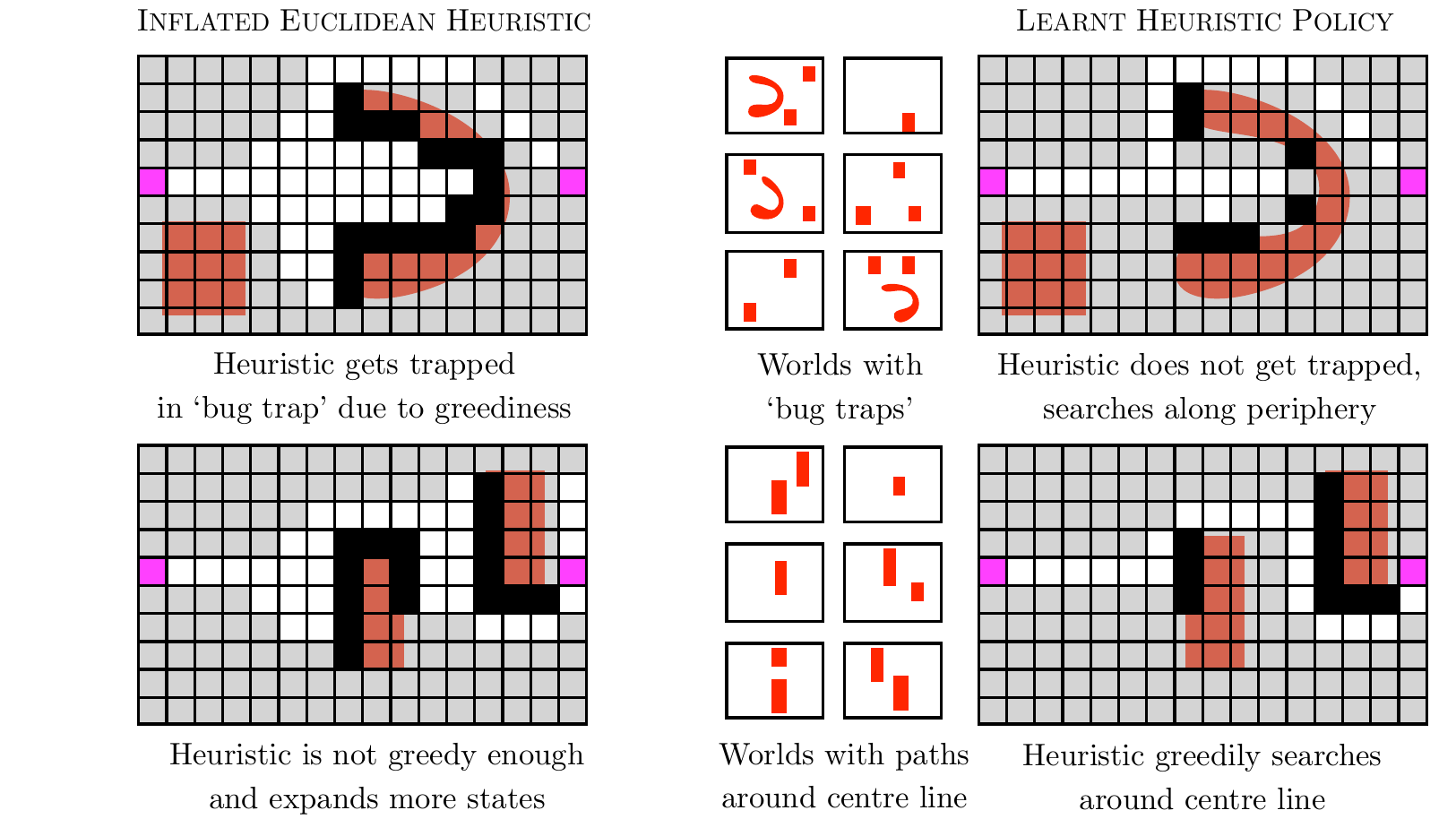}
    \caption{%
    \label{fig:marquee}
    A learnt heuristic policy adapts to different obstacle configurations to minimize search effort. All schematics show the evolution of a search algorithm as the expansion of a search wavefront (expanded(white), invalid(black), unexpanded(grey)) from start to goal (magenta). 
    A commonly used inflated euclidean heuristic cannot adapt to different environments, e.g it gets stuck in bugtraps. 
    On the other hand, the learnt policy is able to infer the presence of a bug trap when trained on such a distribution and switch to greedy behavior when trained on other distributions. \fullFigGap}
\end{figure}

Search based motion planning offers a comprehensive framework for reasoning about a vast number of motion planning algorithms~\citep{Lav06}.
In this framework, an algorithm grows a \emph{search tree} of feasible robot motions from a start configuration towards a goal~\citep{pearl1984heuristics}. 
This is done in an incremental fashion by first selecting a leaf node of the tree, \emph{expanding} this node by computing outgoing edges, checking each edge for validity and finally updating the tree with potentially new leaf nodes.
It is useful to visualize this search process as a \emph{wavefront of expanded nodes} that grows from the start outwards till it finds the goal as illustrated in Fig.~\ref{fig:marquee}. 

This paper addresses a class of robotic motion planning problems where edge evaluation dominates the search effort, such as for robots with complex geometries like robot arms~\citep{dellin2016guided} or for robots with limited onboard computation like UAVs~\citep{cover2013sparse}.
In order to ensure real-time performance, algorithms must prioritize minimizing the search effort, i.e. keeping the volume of the search wavefront as small as possible while it grows towards the goal. 
This is typically achieved by heuristics, which guide the search towards promising areas by selecting which nodes to expand.
As shown in Fig.~\ref{fig:marquee}, this acts as a force stretching the search wavefront towards the goal.

A good heuristic must balance the bi-objective criteria of finding a good solution and minimizing the search effort. 
The bulk of the prior work has focussed on the former objective of guaranteeing that the search returns a near-optimal solution~\citep{pearl1984heuristics}. 
These approaches define a heuristic function as a \emph{distance metric} that estimates the cost-to-go value of a node~\citep{pohl1970first}.
However, estimation of this distance metric is difficult as it's a complex function of robot geometry, dynamics and obstacle configuration. 
Commonly used heuristics such as the euclidean distance do not adapt to different robot configurations or different environments.
On the other hand, by trying to compute a more accurate distance the heuristic should not end up doing more computation than the original search. 
While state of the art methods propose different relaxation-based~\citep{likhachev2009planning, dolgov2008practical} and learning-based approaches~\citep{paden2017verification} to estimate the distance metric they run into a much more fundamental limitation - \emph{a small estimation error can lead to a large search wavefront}. Minimizing the estimation error does not necessarily minimize search effort.  

Instead, we focus on the latter objective of designing heuristics that explicitly reduce search effort in the interest of real-time performance. 
Our key insight is that \emph{heuristics should adapt during search} - as the search progresses, they should actively infer the structure of the valid configuration space, and focus the search on potentially good areas. Moreover, we want to learn this behaviour from data - changing the data distribution should change the heuristic automatically. 
Consider the example shown in Fig.~\ref{fig:marquee}. When a heuristic is trained on a world with `bug traps', it learns to recognize when the search is trapped and circumvent it. On the other hand, when its trained on a world with narrow gaps, it learns a greedy behaviour that drives the search to the goal.

To learn such behaviors, we propose a novel framework for training a data-driven heuristic policy that explicitly minimizes search effort. 
We formulate this as a sequential decision making problem where at a given iteration the heuristic policy uses only the information extracted from the wavefront to decide which node to expand, which in turn influences how the wavefront grows in the next iteration. 
We note that the training process for such partial information based policies is slow to converge and susceptible to local minima. 
We make a useful observation that if the heuristic has full information (which we call \emph{clairvoyant planner}), it can use dynamic programming (Djikstra) to efficiently compute the optimal decisions for any given wavefront. We leverage this property to train heuristics that imitate the clairvoyant planner during train time adopting the framework proposed by \citet{choudhury2017adaptive}. We make the following contributions:
\begin{enumerate}
 \item We propose a novel framework to learn heuristic functions under the paradigm of sequential decision making under uncertainty (Section~\ref{sec:problem_formulation}).
 \item In Section~\ref{sec:algorithm}, we develop \algName, an efficient algorithm for training heuristic functions by imitating clairvoyant oracles.
 \item We demonstrate that we are able to learn heuristic policies with widely varying characteristics simply by training on different data distributions (Section~\ref{subsec:result_analysis}).   
\end{enumerate}
We note that a major limitation of this approach is that it ignores solution quality and we discuss several ways to alleviate this in Section~\ref{sec:conclusion}.

\section{Preliminaries} 
\label{sec:problem_formulation}
\subsection{Search Based Planning Framework} 
\label{subsec:motion_planning_background}
We consider the problem of search on a graph, $\Graph = \pair{\vertexSet}{\edgeSet}$, where vertices $\vertexSet$ represent robot configurations and edges $\edgeSet$ represent potentially valid movements of the robot between these configurations. 
Given a pair of start and goal vertices, $\pair{\start}{\goal} \in \vertexSet$, the objective is to compute a path $\Path \subseteq \edgeSet$ - a connected sequence of valid edges.
The graph $\Graph$ can be defined in one of the following ways: an \emph{explicit graph} which enumerates the edge connections as an adjacency list or an \emph{implicit graph} which is constructed algorithmically during a search.
We use an implicit graph since it's suited for large graphs due to memory efficiency. 
It can be compactly represented by $\pair{\start}{\goal} $ and a successor function $\succFn{\vertex}$ which returns a list of outgoing edges and child vertices for a vertex $\vertex \in \vertexSet$. Hence a graph $\Graph$ is constructed during search by repeatedly \emph{expanding} vertices using $\succFn{\vertex}$.

Let $\world \in \worldSet$ be a representation of the world that is used to ascertain the validity of an edge. 
For example, this could be a map of all obstacles a robot has to avoid. 
An edge $\edge \in \edgeSet$ is checked for validity by invoking an evaluation function $\evalFn(\edge, \world)$. 
For the applications we consider, this is an expensive operation and may require complex geometric intersection operations \citep{dellin2016unifying}.

\begin{algorithm}[!t]
    \caption{$\search \plannerTuple{\start}{\goal}{\succFnDef}{\evalFn}{\world}{\select}$ }\label{alg:search}
    \algorithmStyle
    $\openList \gets \start,\; \closedList \gets \emptyset,\; \closedObsList \gets \emptyset$\;
    \While{$\goal \notin \openList$}
    {
      $\vertex \gets \select(\openList)$ \Comment*[r]{Select a vertex to expand}
      $\pair{\vertexSucc}{\invalidEdge} \gets \expand(\vertex, \succFnDef, \evalFn, \world)$ \Comment*[r]{Invoke $\succFn{\vertex}$ and  $\evalFn(\edge, \world)$}
      $\openList \gets \openList \cup \vertexSucc, \; \closedList \gets \closedList \cup \vertex, \; \closedObsList \gets \closedObsList \cup \invalidEdge $ \Comment*[r]{Update all lists}
    }
    \Return{$\mathtt{Path}\pair{\start}{\goal}$}\;
\end{algorithm}

In this work, we focus on the \emph{feasible path problem}.
Alg.~\ref{alg:search} defines a general search based planning algorithm $\search$ which takes as input the tuple $\plannerTuple{\start}{\goal}{\succFnDef}{\evalFn}{\world}{\select}$ and returns a valid path $\Path$. To ensure systematic search, the algorithm maintains the following lists - an open list $\openList \subset \vertexSet$ of candidate vertices to be expanded and a closed list $\closedList \subset \vertexSet$ of vertices which have already been expanded. It also retains an additional invalid list $\closedObsList \subset \vertexSet$ of edges found to be in collision. These $3$ lists together represent the complete information available to the algorithm at any given point of time. At a given iteration, the algorithm uses this information to select a vertex $\vertex \in \openList$ to expand by invoking $\select(\openList)$. It then expands $\vertex$ by invoking $\succFn{\vertex}$ and checking validity of edges using $\evalFn(\edge, \world)$ to get a set of valid successor vertices $\vertexSucc$ as well as invalid edges $\invalidEdge$. The lists are then updated and the process repeated till the goal vertex $\goal$ is uncovered. 

\subsection{Search as Sequential Decision Making under Uncertainty}
\label{subsec:sequential_decision}
We wish to learn an effective selection strategy $\select$ from data. We formalize this as a problem of sequentially making decisions (selecting vertices) under uncertainty (about the underlying world). We define a corresponding Markov Decision Process (MDP) \footnote{Actually a POMDP which is an MDP over beliefs, referred here as an MDP over states for clarity. } on the space of lists. At timestep $t$, let $\state_t \in \stateSpace$ be the state of the search that is a concatenation of all lists, i.e $\state_t = \{ \openList, \closedList, \closedObsList \}$. The action $\action_t \in \actionSpace$ is the vertex $\vertex \in \openList$ that is selected by the search. On executing $\action_t$, the new state $\state_{t+1}$ is determined by the underlying world $\world$. The world $\world$ is a hidden variable, sampled from a prior $P(\world)$ which in turn induces a state transition distribution $P(\state_{t+1} | \state_t, \action_t)$. The one-step cost $\costonestep\pair{\state_t}{\action_t}$ is defined to be $1$ for every $\pair{\state_t}{\action_t}$ until the goal is added to the open list.
Let $\policy(\state_t)$ be a policy that maps state $\state_t$ to an action $\action_t$. The policy represents the vertex selection strategy that we wish to learn. We term this policy as the \emph{heuristic} guiding the search in a \emph{best-first} fashion towards the goal. An episode continues till either $\goal$ is selected or time horizon $\planTime$ is reached.

Given a prior distribution over worlds $\worldProb$ and a distribution over start and goal vertices $\startProb$, we can evaluate the performance of a policy as 
\begin{equation}
	\label{eq:policy_cost}
	\cost\left( \policy\right) = \expect{
  \substack{\world \sim \worldProb, \\
  \pair{\start}{\goal} \sim \startProb}
  }{\sum_{t=1}^{T}\expect{\state_t\sim\timeAvgDist}{\costonestep\pair{\state_t}{\policy(\state_t)}}}
\end{equation}
where $\timeAvgDist=\prob{\state_t|\policy,\world,\start,\goal}$ is the distribution over states induced by running $\policy$ on the problem $(\world, \start, \goal)$ for $t$ steps \citep{ross2014reinforcement}. Our objective is to learn a policy
\begin{equation}
    \label{eq:optimal_policy}
	\policyOpt = \argminprob{\policy\in\policySpace}\ \cost\left( \policy\right)
\end{equation}

\section{Approach} \label{sec:algorithm}
\subsection{Overview}
An exact solution to the problem in (\ref{eq:optimal_policy}) is intractable given the large state space ($\abs{\stateSpace} = \abs{\vertexSet}$) and complex transition function $P(\state_{t+1} | \state_t, \action_t)$. 
An alternative is to employ model-free Q-learning \cite{mnih2013playing}. These methods try to minimize the cost-to-go $\costToGopolicy{\policy}(\state_t, \action_t)$, i.e, the cumulative cost after executing $\action_t$ from $\state_t$ and subsequently executing policy $\policy$ till the end of the episode. In general, such methods are not very sample efficient, slow to converge and additionally require training strategies such as experience replay and target networks\cite{mnih-dqn-2015}.

However, we leverage a key insight for search based planning problems - if only we had full knowledge of the world $\phi$, we could efficiently compute the optimal action from any state using dynamic programming. While we do not know the world at test time, we know it at train time - we can learn a policy to imitate this action.
Hence we present the \emph{\algFullName (\algName)} algorithm, a simple data-driven imitation learning approach for learning a heuristic best-first search policy. 

\subsection{Imitation Learning with Clairvoyant Oracles} \label{subsec:cost_sensitive_imitation}
Imitation of reference policies (or oracle policies) is a useful approach in scenarios where there exist good reference policies for the original problem, however these policies cannot be executed online (e.g due to computational complexity) hence requiring imitation via an offline training phase. \citet{ross2014reinforcement} use this idea to approach reinforcement learning problems for which there exists good yet expensive \emph{model-based} oracle policies that cannot be executed at run-time but can be imitated by \emph{model-free} policies. Hence they show a novel reduction of reinforcement learning to iterative supervised learning where the labels are the cost-to-go of the oracle. \citet{choudhury2017adaptive} extend this idea to approach POMDP problems (specifically MDPs with a hidden variable) where exists good \emph{clairvoyant oracles} - oracles that could solve the underlying MDP if only they could observe it fully. While such oracles cannot be executed at test time due to an information barrier - they can be imitated in a similar manner.

We adopt the framework of \citet{choudhury2017adaptive} as we too have an MDP whose transition function depends on a hidden world $\world$. We note that we can define an analogous \emph{clairvoyant oracle planner} that employs a \emph{backward} Djikstra's algorithm, which given a world $\phi$ and a goal vertex $\goal$ plans for the optimal path from every $\vertex \in \vertexSet$ using dynamic programming. 
\begin{definition}[Clairvoyant Oracle Planner]
	\label{def:clairvoyant_oracle}
	Given full access to world $\world$ and a goal $\goal$, the oracle planner encodes the cost-to-go from any vertex $\vertex \in \vertexSet$ as the function $\costToGoOracle\pair{\vertex}{\world}$ which implicitly defines an oracle policy, $\policyOracle(\state, \world) \; = \; \argminprob{\vertex\in\openList} \; \costToGoOracle\pair{\vertex}{\world}$.
\end{definition}
The clairvoyant oracle planner provides a look-up table $\costToGoOracle\pair{\vertex}{\world}$ for the optimal cost-to-go from any vertex irrespective of the current state of the search. 
We define imitation of such an oracle as the following cost sensitive classification:
\begin{equation}
\label{eq:learnt_policy}
\policyLearn\left(\state\right) = \argminprob{\policy\in\policySpace} \;\expect{\substack{\world\sim\worldProb
		\\t \sim \uniform{1 \ldots \planTime}
	    \\\state\sim\timeAvgDist}}{\costToGoOracle 
	\pair{\policy\left(\state \right) }{\world}\; 
	- \ \argminprob{\vertex \in \openList} \ \costToGoOracle\pair{\vertex}{\world}}
\end{equation}
Intuitively, the term inside the expectation in Eq. \ref{eq:learnt_policy} scores the learner's mis-classifications (incorrect vertex expansions) by how much additional future cost the oracle would incur if it chose the same action instead of following its own policy. Given a world $\world \sim \worldProb$, if a policy $\policy$ is executed upto a uniformly sampled timestep \emph{t}, this scoring metric implicitly induces a ranking among all the states in the resulting open-list. 

In order to learn the above policy, we use a reduction of cost-sensitive classification to $argmin$ \emph{regression}. Our aim is to learn a parameterized function $\costToGo_{\thetaLearn}\left(\state_t, \action_t\right)$ that takes the current state and action as input and approximates $\costToGoOracle 
\pair{\vertex}{\world}$, where $\action_t$ is $\expand(\vertex, \succFnDef, \evalFn, \world)$. Using the learnt parameters $\thetaLearn \in \thetaSpace$, the planner follows a greedy policy given by, 
\begin{equation}
	\policyLearn(\state_t) = \argminprob{\action_t \in \actionSpace}\; \costToGo_{\thetaLearn}\left(\state_t, \action_t \right)
\end{equation}
where $\thetaLearn$ is learnt using the following procedure
\begin{equation}
    \label{eq:theta_learn}
	\thetaLearn =  \argminprob{\theta \in \thetaSpace}\;
	\expect{\substack{\world\sim\worldProb
			\\ t \sim \uniform{1 \ldots \planTime}
			\\\state\sim\timeAvgDist}}{\left(\costToGo_{\theta}\left(\state_t, \action_t \right) \; - \; \costToGoOracle\pair{\vertex}{\world}\right)^{2}}
\end{equation}
It is important to note that data for the learning the policy (Eq. \ref{eq:learnt_policy} and Eq. \ref{eq:theta_learn}) needs to be collected on the true distribution of states, $\timeAvgDist$ induced by executing policy $\policy$ on world $\phi$.
A key distinction between our framework and that of \citet{choudhury2017adaptive} is that we directly get the cost-to-go value for all states by dynamic programming - we do not need to repeatedly invoke the oracle. We exploit this fact by extracting multiple labels from an episode even though the oracle is invoked only once.

\subsection{\algName Algorithm} \label{subsec:sail_algo}

\begin{algorithm}[t]
	\caption{\algName}\label{alg:sail_alg}
	Initialize $\dataset \leftarrow \emptyset,\; \policyLearn_{1}$ to any policy in $\policySpace$ \\ \label{lst:line:}
	\For{$i = 1, \ldots, N$}
	{   Initialize sub-dataset $\dataset_{i} \leftarrow \emptyset$ \\
		Let mixture policy be $\policyMix =\;\mixParam_{i}\policyOracle\; + \; (1 \; - \;\mixParam_{i})\policyLearn_{i} $ \\
		Collect \emph{mk} datapoints as follows:\\
		\For{$j = 1,\ldots,m$}
		{   
			Sample $\world \sim P(\world)$; \\  
			Sample $\pair{\start}{\goal} \sim \startProb$; \\
			Invoke clairvoyant oracle planner to compute $\costToGoOracle\pair{\vertex}{\world} \forall \vertex \in \vertexSet$; \\
			Sample uniformly \emph{k} timesteps $\seqset{t}{k}$ where each $t_{i} \in \ \set{1, \ldots ,\planTime}$;\\
			Rollout a new search with $\policyMix$\Comment*[r]{Invoke Alg.\ref{alg:search} with $\select$ as $\policyMix$}  
			At each $t\in\seqset{t}{k}$ pick a random action $\action_t$ to get corresponding $\pair{\vertex}{\state_t}$;\\
			Query oracle for $\costToGoOracle\pair{\vertex}{\world}$ \Comment*[r]{Look-up optimal cost-to-go}
		    $\dataset_i \gets \dataset_{i} \cup \left< \vertex, \state_t, \costToGoOracle\pair{\vertex}{\world} \right>$ \Comment*[r]{Collect data for Eq.\ref{eq:theta_learn}}
		    Continue roll-out with $\policyMix$ till end of episode.;
	    }
	    Append to c.s classification data $\dataset \leftarrow \dataset \cup \dataset_{i}$; \\
	    Train $\thetaLearn_{i+1}$  on $\dataset$ to get $\policyLearn_{i+1}$;\\  
	}
	\Return{Best $\policyLearn$ on validation};
\end{algorithm}

Alg.~\ref{alg:sail_alg}, describes the $\algName$ framework which iteratively trains a sequence of policies $\seq{\policyLearn}{N}$. For the optimization procedure described in Eq. \ref{eq:theta_learn}, we collect a dataset $\dataset$ as follows - At every iteration \emph{i}, the agent executed \emph{m} different searches (Alg. \ref{alg:search}). For every search, a different world $\world$ and the pair $(\start, \goal)$ is sampled from a database.  The agent then rolls-out a search with a mixture policy $\policyMix$ which blends the learner's current policy, $\policyLearn_{i}$ and the oracle's policy, $\policyOracle$ using blending parameter $\mixParam_{i}$. During the search execution, at every timestep in a set of $k$ uniformly sampled timesteps, we select a random action from the set of feasible actions and collect a datapoint $<\vertex, \state_t, \costToGoOracle(\vertex, \world)>$. The policy $\policyMix$ is rolled out till the end of the episode and all the collected data is aggregated with dataset $\dataset$. The optimization in Eq. \ref{eq:theta_learn} can then be performed using either \emph{online} or \emph{mini-batch} learning on $\dataset$ to get the next policy $\policyLearn_{i+1}$.
\begin{equation}
\label{eq:theta_learn_dataset}
\thetaLearn_{i+1} =  \argminprob{\theta \in \thetaSpace}\;
\expect{\substack{(\state_t, \action_t, \costToGoOracle) \sim \dataset}}{\left(\costToGo_{\theta}\left(\state_t, \action_t \right) \; - \; \costToGoOracle\pair{\vertex}{\world}\right)^{2}}
\end{equation}
At the end of N iterations, the algorithm returns the best performing policy on a set of held-out validation environment or alternatively, a mixture of $\seq{\policyLearn}{N}$. Note that while the oracle is invoked once per $\world$, we obtain $k$ datapoints - this is critical for speeding up training.

We can obtain performance guarantees on the learnt policy directly applying analysis from~\citep{choudhury2017adaptive}
\begin{theorem}
The performance of the returned policy from Alg.~\ref{alg:sail_alg} is, with probability at least $1-\delta$
\begin{equation*}
\begin{aligned}
  J(\policyLearn) \leq & J(\policyOracleBel) + 2\sqrt{\abs{\actionSpace}}T\sqrt{\errclass + \errreg + O \left(\sqrt{\log \left(\nicefrac{ \left( \nicefrac{1}{\delta} \right)}{Nm}\right)} \right)} + O \left(\frac{\costToGoOracle_{\mathrm{max}}T \log T}{\alpha N}\right)
\end{aligned}
\end{equation*}
\end{theorem}

\section{Experiments} 
\label{sec:experiments}
\subsection{Implementation Details} \label{subsec:implementation}
We evaluate \algName on a variety of 2D navigation tasks where the robot has to plan from bottom-left to top-right on an 8-connected grid. The grid is embedded on a binary map of obstacles. 
The function $\costToGo_{\theta}$ is represented by feed-forward neural network with two fully connected hidden layers containing [100, 50] units and ReLu activation. The model takes as input a 17 dimensional feature vector $\featureVec \in \featureSpace$ for the pair $(\vertex, \state)$ which contains values like closest invalid state in $\closedObsList$, distance to start and goal, depth in the tree etc. 
Refer to \hyperref[subsec:feature_representation]{supplementary} for details.\footnote{Open source \hyperref[subsec:experimental_setup]{code} for planning pipeline, an OpenAI Gym~\citep{brockman2016gym} environment and datasets will be provided.}

\subsection{Baseline Approaches For Heuristic Search} \label{subsec:baselines}
\textbf{Motion Planning Baselines}: We compare against greedy best-first search with 2 commonly used heuristics - the euclidean distance ($\greedyEuc$) and the manhattan distance ($\greedyMan$). We also use A* algorithm as a baseline with $\greedyEuc$ heuristic. Additionally, we compare against the MHA* algorithm~\citep{aine2016multi} which has been proven to be an effective way of combining multiple, often unrelated, heuristics providing bounds on solution quality~\citep{narayanan2015improved, islam2015dynamic, phillips2015efficient}. We use a simplified version which expands three different heuristics in a round-robin fashion - $\left[\greedyEuc, \greedyMan, \distObs\right]$, where $\distObs$ is the euclidean distance to closest, \emph{known} obstacle cell in $\closedObsList$.

\textbf{Machine Learning Baselines}: We consider two learning baselines (a) Supervised Learning (SL) with data from roll-outs with $\policyOracle$ and (b) Reinforcement Learning using evolutionary strategies (CEM) and Q-Learning (QL) with function approximation. Refer to \hyperref[subsec:baseline_details]{supplementary} for details.

\begin{figure}[!t]
	\centering
	\includegraphics[page=1,width=\textwidth]{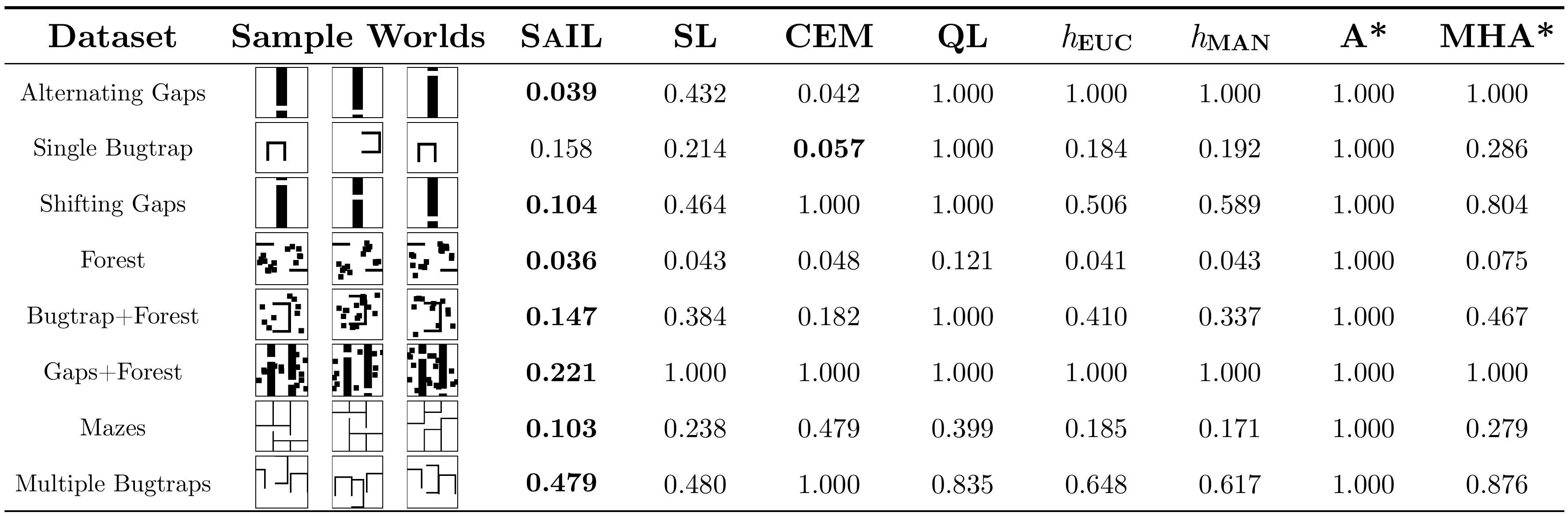}
	
	\caption{\label{fig:results_table} 
		Normalized cost of baselines on different environments (best in bold). 
		Planning parameters are - map size: $200\times200$,$\planTime_{train}=1100$, $\planTime_{test}=20000$. 
		Data sizes are: train($200$), test($100$), validation($70$). 
		\algName parameters are - $\dataPointsParam: 50, \beta_{0} = 0.7$.
		\algName, CEM and QL are run for $N: 15$ iterations.
		SL uses $m:600$. \fullFigGap}
\end{figure}%

\subsection{Analysis of Results}
\label{subsec:result_analysis}
\begin{figure}[!t]
	\centering
	\includegraphics[page=1,width=\textwidth]{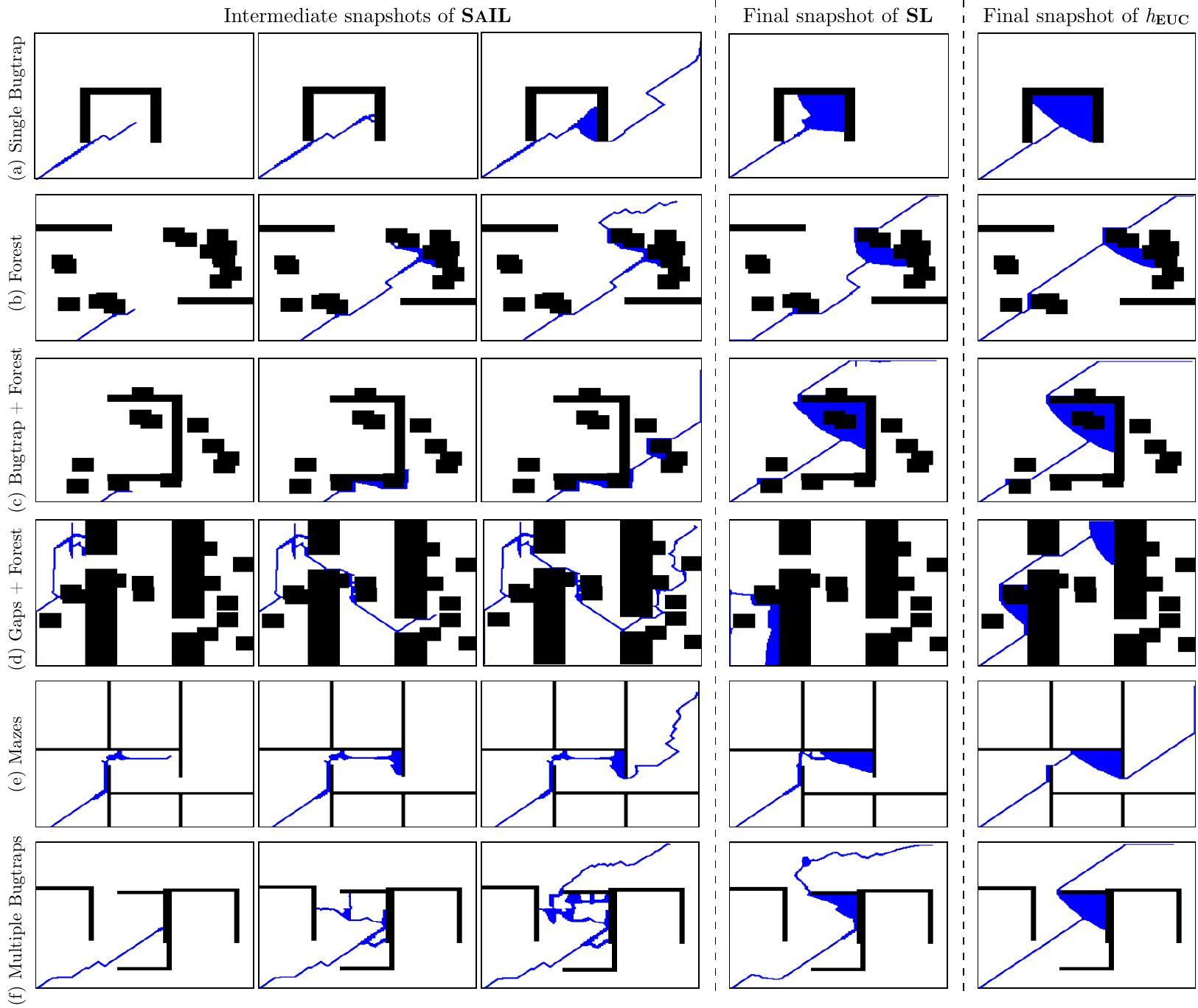}
	\caption{\label{fig:benchmark_results}
	Evolution of search frontier (expanded(blue), invalid(black), unexpanded(white)) of \algName compared with final snapshot of supervised learning (SL) and $\greedyEuc$ across all environments. \algName expands far less states. \fullFigGap}
\end{figure}

\begin{figure}[!t]
	\centering
	\includegraphics[page=1,width=\textwidth]{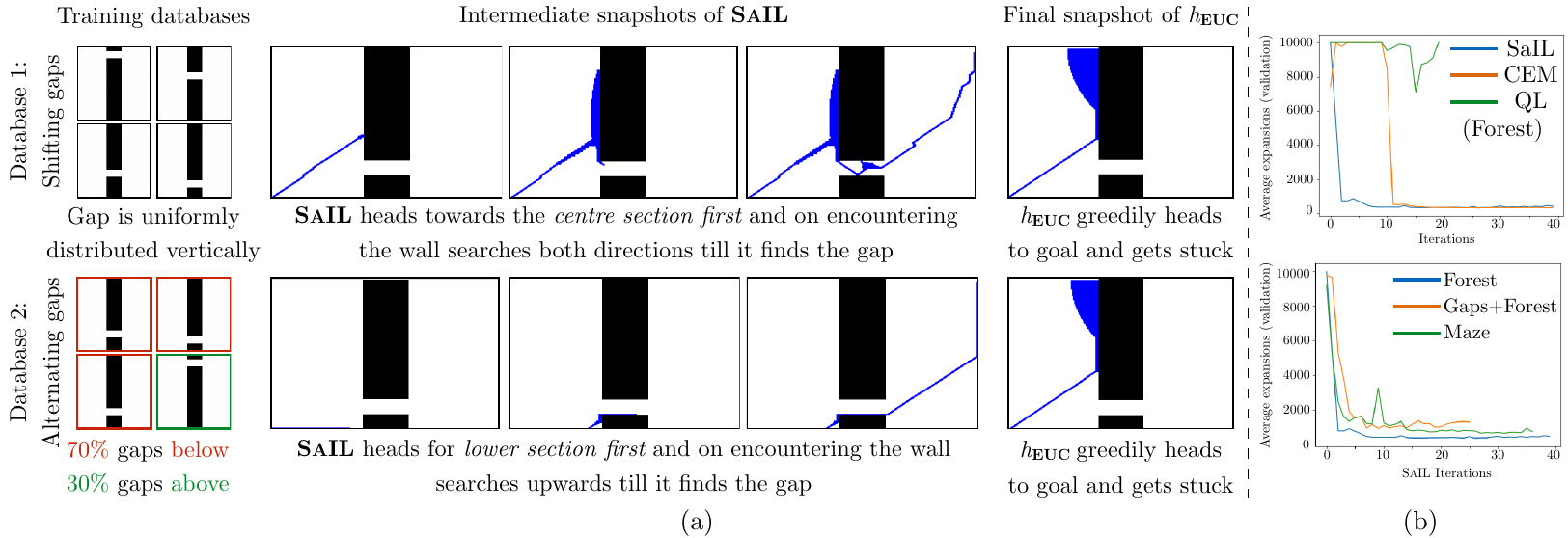}
	\caption{\label{fig:explanatory_result}
	(a) \algName learns to adapt to different environment distributions by directing search to areas where it expects to find gaps.
	 Note \algName does not have information about the entire environment, only the explored part. 
	 (b) On the `Forest' dataset, \algName converges faster that CEM and QL to a good policy. \algName also converges consistently to a good policy across environments `Gaps', `Gaps+Forest', `Maze.' \fullFigGap}
\end{figure}

Fig.~\ref{fig:results_table} shows the normalized evaluation cost of all algorithms on various datasets. 
Snapshots of planning with different heuristics are shown in Fig.~\ref{fig:benchmark_results} and Fig.~\ref{fig:explanatory_result}(a). 
Convergence of different learning algorithms are shown in Fig.~\ref{fig:explanatory_result}(b).
We present a set of observations to interpret these results.

\begin{observation}
\algName has a consistently competitive performance across all datasets.
\end{observation}\vspace{-0.7em}

Fig.~\ref{fig:results_table} shows that \algName learns a better search policy than any other baseline across all but one environments. It maintains performance from homogenous to heterogenous environments. 

\begin{observation}
\algName has faster convergence than all learning baselines.
\end{observation}\vspace{-0.7em}

Fig.~\ref{fig:explanatory_result}(b) shows that on the `Forest' dataset, \algName converges by $6^\text{th}$ iteration, while CEM takes $12$ and QL does not converge. \algName also converges quickly (by the $8^\text{th}$ iteration) across datasets.

\begin{observation}
	\algName adapts the behavior of the search in response to a change in world distribution $\worldProb$.
\end{observation}\vspace{-0.7em}

Fig.~\ref{fig:explanatory_result}(a) shows an intuitive example of how simply changing the distribution of where gaps occur along a wall affects the way \algName chooses to progress the search.

\begin{observation}
	\algName is able to detect and escape local minima.
\end{observation}\vspace{-0.7em}

A classic case in motion planning is the bugtrap (Fig.~\ref{fig:marquee}) which traps a greedy search in a local minimum. Fig.~\ref{fig:benchmark_results}(a) and Fig.~\ref{fig:benchmark_results}(f) shows that when trained on such distributions, \algName is able to detect these artifacts and smartly escape them by exploring in different directions.

\begin{observation}
	\algName is able to exploit the relative configuration of obstacles and environment structure. 
\end{observation}\vspace{-0.7em}

In a maze world with rectilinear hallways (Fig.~\ref{fig:benchmark_results}(e)), \algName learns to quickly find a wall and then concentrate the search along the axes. In Fig.~\ref{fig:benchmark_results}(d), \algName focuses only on regions where there is a high probability of a gap and skids along obstacles otherwise.

\section{Related Work} \vspace{-0.3em}
\label{sec:related_work}
Learning heuristics falls under machine learning for general purpose planning~\citep{jimenez2012review}. \citet{yoon2006learning}~\citep{yoon2008learning} propose using regression to learn residuals over FF-Heuristic~\citep{hoffmann2001ff}. \citet{xu2007discriminative}~\citep{xu2010iterative, xu2009learning} improve upon this in a beam-search framework. \citet{arfaee2011learning} iteratively improve heuristics. \citet{us2013learning} learn combination of heuristic to estimate cost-to-go. Kendall rank coefficient is used to learn open list ranking~\citep{wilt2015building,garrettlearning}. \citet{thayer2011learning} learn heuristics online during search. \citet{paden2017verification} learn admissible heuristics as S.O.S problems. However, these methods do not address minimization of search effort and also ignore the non i.i.d nature of the problem.

Relevant work in imitation learning examines the non i.i.d supervised learning problem of imitating oracles under one's own state distribution. \citet{ross2011reduction,ross2014reinforcement} use dataset aggregation to reduce such problems to no-regret iterative supervised learning.
\citet{choudhury2017adaptive, Choudhury_2017_22858} apply such methods to learn information gathering policies.
Recent deep reinforcement learning approaches also employ supervised learning by imitating oracles as they offer better sample efficiency and safety than model free policy search~\citep{schulman2015trpo, schulman2015gae,lillicrap2015ddpg} or Q learning~\citep{mnih-dqn-2015,wang2016dueling}. \citet{zhang2016mpcgps} extend guided policy search~\citep{levine2013guided} for imitating MPC. \citet{kahn2016plato} further adapt the MPC expert to generate training sample for states likely to be visited. \citet{tamar2016hindsight} consider a cost-shaping approach for short horizon MPC by offline imitation of long horizon MPC which is closest to our work. \citet{tamar2016vin} develop a neural network architecture with an explicit planning component embedded in it. \citet{gupta2017cmp} develop a holistic mapping and planner framework trained using feedback from optimal plans on a graph. 
\vspace{-0.7em}\section{Discussion and Future Work}\vspace{-0.3em}
\label{sec:conclusion}
In this paper, we addressed the problem of learning a heuristic policy that guides a search based planner to find a feasible solution 
while minimizing search effort.
We showed this problem can be formulated as a sequential decision making under uncertainty problem.
We proposed \algName, and efficient approach to training heuristic policies by imitating a value function at train time. 
We validated \algName on a spectrum of problems against state of the art baselines and shows that it consistently outperforms them.
We now discuss some insights and directions for future work.
\begin{question}
	When do we expect this framework of imitating clairvoyant oracles to work?
\end{question}\vspace{-0.5em}
The analysis adopted from \citep{choudhury2017adaptive} states that the performance of the learnt policy using \algName is near-optimal with respect to a \emph{hallucinating oracle} - an oracle that hallucinates different worlds conditioned on the current open list and expands the best node. The hallucinating oracle is similar in nature to a QMDP algorithm~\citep{Littman95learningpolicies}, an effective approximate solution to POMDPs, which takes the best action on the current posterior. However, while QMDP is model based (requires an explicit posterior), \algName is model free. QMDP has been shown to be very successful where explicit information gathering behavior is not required~\citep{Koval-RSS-14,javdani2015shared} - the belief collapses irrespective of the action. This is very apt in the problem we address - as the set of actions are constrained to candidate nodes in the open list, no single action is very informative. It suffices to expand the best node under the current belief and continue to update the belief as the open list evolves. We note that this is not true for all learning in planning paradigms. For example, when learning to collision check~\citep{choudhury2017active}, a policy that actively reduces uncertainty about the world is effective. This is because the policy, in this case, is free to check any edge in the graph and has more chance to explore. A direction for future work would be to have a more unified framework that encompasses both types of problems.

\begin{question}
	How can we incorporate solution cost in addition to search effort in this framework?
\end{question}\vspace{-0.5em}
While our framework ignores the cost of a solution, we note that finding feasible solutions quickly is the core motivation of a number of high-dimensional planning problems which have historically resorted to sampling based approaches~\citep{kuffner2000rrt}. 
Hence, one can apply our framework to such problems to produce potentially faster solutions.
We also note that when planning on locally connected lattices for geometric planning problems, minimizing the number of expansions generally leads to near-optimal solutions (unit cost for each valid edge). 
However, if we really cared about near optimal solutions, the framework of Multi-heuristic A* (MHA*)~\citep{aine2016multi} can be easily adopted. In such a framework, any heuristic function~\citep{narayanan2015improved} can be used in tandem with an anchored search which uses an inflated admissible heuristic. Hence we can simply replace our $\search$ function with MHA*. 
The bi-objective criteria of solution cost and search effort is best reasoned about in the paradigm of \emph{anytime planning}. In this paradigm, an algorithm traces out the \emph{pareto-froniter}~\citep{choudhury2016pareto} - finds a feasible solution quickly and iteratively improves it. In this paradigm, \algName trains a heuristic that displays a behavior we would expect in the first iteration. A direction of future work would be to learn \emph{anytime heuristics} that minimize search effort initially to and solution cost eventually.
%
\bibliographystyle{plainnat}
{\small
\bibliography{reference}
}
\newpage
\section{Supplementary Material} \label{sec:supplementary}
\subsection{Experimental Setup}
\label{subsec:experimental_setup}
The initial phase of experimentation was carried out on a custom GraphSearch environment for 2D navigation planning compatible with OpenAI Gym~\citep{brockman2016gym}. The intention behind the development of this environment was to easily benchmark the performance of state-of-the-art Reinforcement Learning and Imitation Learning algorithms on learning search heuristics. 

In order to overcome the changing sizes of the observation and action spaces in our setting, we use insight from motion planning literature and represent an entire search state in terms of closest nodes in $\openList$ to a set of pre-defined \emph{attractor states} and \emph{attractor paths}. Attractor states are manually defined states that can be thought of as landmarks trying to pull the search cloud in different directions. Such states can be useful in pulling the search out of local minima such as a bugtrap or they could be strategic orientations of the robot or an object the robot is trying to manipulate that lead to faster solutions~\citep{aine2016multi}. Attractor paths, on the other hand, are solutions to a small subset of environments from the training dataset. In many episodic tasks, where the structure of the environment does not change drastically between planning iterations, such \emph{path-reuse} can be very useful in finding solutions faster~\citep{phillips2012graphs}. The planning algorithm is built into the environment, and the agent only receives as an observation the nodes in the open list closest to each attractor paths/states. At each iteration then, the action that the agent performs is to select a node from the observation to expand.  

Although this is a generic framework that can be applied to many different problems, we chose not to use it for this work. The reason for this choice was that in this paper, our aim was to build the foundation for learning graph search heuristics as a sequential decision making problem and clearly demonstrate the efficacy of the imitation learning paradigm in this domain. We found that using attractor paths/states would distract from the effectiveness of $\algName$ and also make learning easier for other baseline methods. As future work, we wish to create more concrete problems for the GraphSearch environment and benchmark the performance of state-of-the-art algorithms. 

In our final experiments, we instead featurize every pair $\pair{\vertex}{\state}$ using simple information based on the search tree and the environment uncovered up until that point. We talk more about the features in a later subsection. Additionally, we have developed a simple and intuitive Python based planning pipeline to serve as a backend for the GraphSearch environment. The planning environment makes it easy to incorporate machine learning libraries \cite{2016arXiv160502688short} \cite{DBLP:journals/corr/AbadiABBCCCDDDG16} with custom planning graphs requires only environment images as input. We use this planning pipeline to conduct all our experiments. Code for the planning pipeline can be found \href{https://goo.gl/StPEmV}{here} and the Gym environment can be found \href{https://goo.gl/Lifjin}{here}.
\subsection{Feature Representation}
\label{subsec:feature_representation}
We explore different ways to construct representative features for a node in the search tree. Although technically, the features for a vertex $\vertex$ should depend on the parent edge $\edge$ that leads to the vertex, we ignore this in practice and consider a vertex in isolation to calculate features. 
It is important to note that the features used must be easy to calculate (no high computational burden) and should only require information uncovered by search until that point in time(else it would count as extra expansions). To this end we divide our feature into two categories as follows:

\emph{Search Based Features:} $\featureVec_{S}\pair{\vertex}{\state}$ = $\left[x_{\vertex}, y_{\vertex}, \gVal, \greedyEuc, \greedyMan, \treeDepth, x_{\goal}, y_{\goal} \right]$, where:
\begin{enumerate}
	\item[-] $(x_{\vertex}, y_{\vertex})$ - location of node in coordinate axis of occupancy map. 
	\item[-] $(x_{\goal}, y_{\goal})$ - location of goal in coordinate axis of occupancy map.
	\item[-] $\gVal$ - distance to start.
	\item[-] $\greedyEuc$ - Euclidean distance to goal.
	\item[-] $\greedyMan$ - Euclidean distance to goal.
	\item[-] $\treeDepth$ - Depth of $\vertex$ in the search tree so far.
\end{enumerate}
\emph{Environment Based Features:} These features depend upon the environment uncovered so far.
$\featureVec_{E}\pair{\vertex}{\state}$ = $\left[x_{OBS}, y_{OBS}, d_{OBSX},x_{OBSX}, y_{OBSX}, d_{OBSX}, x_{COBSY}, y_{OBSY}, d_{OBSY} \right]$, where:
\begin{enumerate}
	\item[-] $x_{OBS}, y_{OBS}, d_{OBSX}$ - coordinates and distance of closest node in $\closedObsList$ to $\vertex$
	\item[-] $x_{OBSX}, y_{OBSX}, d_{OBSX}$ - coordinates and distance of closest node in $\closedObsList$ to $\vertex$ in terms of  x-coordinate. 
	\item[-] $x_{OBSY}, y_{OBSY}, d_{OBSY}$ - coordinates and distance of closest node in $\closedObsList$ to $\vertex$ in terms of y-coordinate
\end{enumerate} 
The net feature vector is then a concatenation of the two vectors i.e, $\featureVec = \left[\featureVec_{S}, \featureVec_{E}\right]$ which gives a 17-dimensional feature vector. 

We also explored the possibility of using a local image patch representing the explored obstacles in black and free space in white as $\featureVec_{E}$, but found that the patch used must be large enough to contain useful information. Testing the efficacy of the image patch as a feature and combining it with different machine learning models (eg. CNNs) would require further experimentation especially the computational	 trade-offs it poses.  

\subsection{Network Architecture and Hyperparameters}
\label{subsec:network}
The function $\costToGo_{\theta}$ is represented using a feed-forward neural network with two fully connected hidden layers containing [100, 50] units and ReLu activation. The model takes as input a feature vector $\featureVec \in \featureSpace$ for the pair $(\vertex, \state)$ and outputs a scalar cost-to-go estimate. The network is optimized using RMSProp \cite{rmsprop}. A mini-batch size of 64 and a base learning rate of 0.01 is used. The network architecture and hyper-parameters are kept constant across all environments. 

\subsection{Practical Algorithm Implementation}
\label{subsec:practical_algorithm}
Since the size of the action space changes as more states are expanded, the \algName algorithm requires a forward pass through the model for every action individually unlike the usual practice of using a network that outputs cost-to-go estimate for all actions in one pass as in \cite{mnih-dqn-2015}. This can get computationally demanding as the search progresses ($\bigo{N}$  in actions). Instead, we use a \emph{priority queue} as $\openList$ which sorts vertices in increasing order of the cost-to-go estimates as is usually done in search based motion planning. The vertex on the top of the list is then expanded. We use two priority queues, sorted by the learner and oracle's cost-to-go estimates respectively. This allows us to take actions in $\bigo{1}$ but forces us to freeze the \emph{Q}-value for a vertex to whenever it is inserted in $\openList$. Despite this artificial restriction over the policy class $\policySpace$, we are able to learn efficient policies in practice. However, we wish to relax this requirement in future work.

\subsection{Details of Baseline Algorithms}
\label{subsec:baseline_details}
\textbf{Supervised Learning (Behavior Cloning)} \\
The supervised learning algorithm is identical to $\algName$ with the key difference that roll-outs are made with $\policyOracle$ and not $\policyMix$. This is equivalent to setting the mixing parameter $\mixParam = 1$ across all environments. For completeness, we present the algorithm below in Alg.~\ref{alg:supervised_learning}
\begin{algorithm}[t]
	\caption{Supervised Learning }\label{alg:supervised_learning}
	Initialize $\dataset \leftarrow \emptyset$, \\ \label{lst:line:}
	Collect datapoints as follows:\\
	\For{$i = 1, \ldots, m$}
	{   Initialize sub-dataset $\dataset_{i} \leftarrow \emptyset$; \\
		
		Sample $\world \sim P(\world)$; \\  
		Sample $\pair{\start}{\goal} \sim \startProb$; \\
		Invoke clairvoyant oracle planner to compute $\costToGoOracle\pair{\vertex}{\world} \forall \vertex \in \vertexSet$; \\
		Rollout a new search with $\policyOracle$;\\ 
		At each timestep $t$ pick a random action $\action_t$ to get corresponding $\pair{\vertex}{\state_t}$;\\
		Query oracle for $\costToGoOracle\pair{\vertex}{\world}$ \Comment*[r]{Look-up optimal cost-to-go}
		$\dataset_i \gets \dataset_{i} \cup \left< \vertex, \state_t, \costToGoOracle\pair{\vertex}{\world} \right>$; \\
		Continue roll-out with $\policyOracle$ till end of episode.;
	}
	Append to c.s classification data $\dataset \leftarrow \dataset \cup \dataset_{i}$; \\ 
	Train $\thetaLearn$  on $\dataset$ to get $\policyLearn$;\\ 
	\Return $\policyLearn$
\end{algorithm}
We use $m = 600$ for all the environments. The network architecture and hyper-parameters used are the same as described in the prior subsection.

\textbf{Episodic Q-Learning} \\
We developed an episodic implementation of the Q-learning algorithm which collects data in an iteration-wise manner similar to \algName. $\costToGo_{\theta}$ is then trained on the aggregated dataset across all iterations by regressing to the TD-error. The aggregated dataset $\dataset$ effectively acts as an experience replay buffer which helps stabilize learning when using neural network function approximation as has been suggested in recent work~\citep{mnih-dqn-2015}. However, we do not use a target network or any other extensions over the original Q-learning algorithm in our baselines, ~\citep{DBLP:journals/corr/HasseltGS15, DBLP:journals/corr/SchaulQAS15}. We also use only a single observation to take decisions and not a history length of past \emph{h} observations for a fair comparison with \algName which also uses a single observation. Alg.~\ref{alg:q_learning} describes the training procedure for the Q-learning baseline.
\begin{algorithm}[t]
	\caption{Episodic Q-Learning}\label{alg:q_learning}
	Initialize $\dataset \leftarrow \emptyset,\; \policyLearn_{1}$ to any policy in $\policySpace$ \\ \label{lst:line:}
	\For{$i = 1, \ldots, N$}
	{   Initialize sub-dataset $\dataset_{i} \leftarrow \emptyset$ \\
		Let mixture policy be $\policyMix = \epsilon  \text{-greedy on} \;  \policyLearn_{i} \; \text{with} \; \epsilon_{i} $ \\
		Collect \emph{mk} datapoints as follows:\\
		\For{$j = 1,\ldots,m$}
		{   
			Sample $\world \sim P(\world)$; \\  
			Sample $\pair{\start}{\goal} \sim \startProb$; \\
			Sample uniformly \emph{k} timesteps $\seqset{t}{k}$ where each $t_{i} \in \ \set{1, \ldots ,\planTime}$;\\
			Rollout a new search with $\policyMix$;\\
			At each $t\in\seqset{t}{k}$, 
			$\dataset_i \gets \dataset_{i} \cup \left< \vertex, \state_t, C, \vertex_{t+1} \right>$ \Comment*[l]{$\vertex_{t+1}$ is the least-Q action after executing $a_t$ in $s_t$}
			Continue roll-out with $\policyMix$ till end of episode.;
		}
		Append to dataset $\dataset \leftarrow \dataset \cup \dataset_{i}$; \\
		Train $\thetaLearn_{i+1}$ by minimizing T.D error  on $\dataset$ to get $\policyLearn_{i+1}$;\\  
	}
	\Return{Best $\policyLearn$ on validation};
\end{algorithm}
C is the one step cost which is 1 for every expansion till goal is added to the open list. We use $k = 100$ and $\epsilon_{0} = 0.9$. Epsilon is decayed after every iteration in an exponential manner. Network architecture and params are kept the same as $\algName$.

\textbf{Cross Entropy Method (C.E.M)}
We use C.E.M as a derivative free optimization method for training $\costToGo_{\thetaLearn}$. At each iteration of the algorithm we sample size batch\_size = 40 set of parameters from a Gaussian Distribution. Each parameter is used to roll-out a policy on 5 environments each and the total cost is collected. The total cost (number of expansions) is used as the fitness function and the 
the best performing, $n_{elite} = 20\%$ of the parameters are selected. These elite parameters are then used to create a new Gaussian distribution (using sample mean and standard deviation) for the next iteration. At the end of all iterations, the best performing policy on a set of held-out states is returned. For this baseline, we use a simpler neural network architecture with one hidden layer of 100 units and ReLu activation.

\end{document}